\documentclass[10pt,twocolumn,letterpaper]{article}

\usepackage{cvpr}              
\usepackage{times}
\usepackage{epsfig}
\usepackage{graphicx}
\usepackage{amsmath}
\usepackage{amssymb}
\usepackage{paralist}

\usepackage{booktabs}
\usepackage{multicol}
\usepackage{multirow}
\usepackage{bm}
\usepackage{bbm}
\usepackage{diagbox}
\usepackage{makecell}

\usepackage{color, colortbl}
\newcommand{\gr}{\rowcolor[gray]{.95}} 

\usepackage[dvipsnames]{xcolor}

\definecolor{cvprblue}{rgb}{0.21,0.49,0.74}
\usepackage[pagebackref,breaklinks,colorlinks,citecolor=cvprblue]{hyperref}

\def\ours{RCBEVDet}
\newcommand{\figref}[1]{Figure~\ref{#1}}%
\newcommand{\tabref}[1]{Table~\ref{#1}}%
\newcommand{\secref}[1]{Section~\ref{#1}}
\renewcommand{\eqref}[1]{Eq.~(\ref{#1})}

\def\paperID{3660} 
\def\confName{CVPR}
\def\confYear{2024}

\title{RCBEVDet: Radar-camera Fusion in Bird's Eye View for 3D Object Detection}

\author{
Zhiwei Lin$^{1}$\footnotemark[1]\quad  
Zhe Liu$^{2}$\footnotemark[1]\quad  
Zhongyu Xia$^1$\quad 
Xinhao Wang$^{1}$\quad
Yongtao Wang$^{1}$\footnotemark[2]\quad\\
Shengxiang Qi$^{3}$\quad
Yang Dong$^{3}$\quad
Nan Dong$^{3}$\quad
Le Zhang$^{2}$\quad
Ce Zhu$^{2}$\\
$^1$Wangxuan Institute of Computer Technology, Peking University \quad
$^2$School of Information \\ and Communication Engineering, 
University of Electronic Science and Technology of China \\
$^3$Chongqing Changan Automobile Co., Ltd.\\
{\tt\small \{zwlin,wyt,xiazhongyu\}@pku.edu.cn\quad liuzhe@std.uestc.edu.cn \quad shengxiang.qi@gmail.com} \\ 
{\tt\small \{lezhang,eczhu\}@uestc.edu.cn\quad \{dongyang,dongnan1\}@changan.com.cn }
}

\begin{document}
\maketitle
\renewcommand{\thefootnote}{\fnsymbol{footnote}}
\footnotetext[1]{Equal contribution.\quad\quad\quad $^{\dagger}$Corresponding author.} 


\begin{abstract}
Three-dimensional object detection is one of the key tasks in autonomous driving.
To reduce costs in practice, low-cost multi-view cameras for 3D object detection are proposed to replace the expansive LiDAR sensors.
However, relying solely on cameras is difficult to achieve highly accurate and robust 3D object detection.
An effective solution to this issue is combining multi-view cameras with the economical millimeter-wave radar sensor to achieve more reliable multi-modal 3D object detection.
In this paper, we introduce \ours, a radar-camera fusion 3D object detection method in the bird's eye view (BEV).
Specifically, we first design RadarBEVNet for radar BEV feature extraction. 
RadarBEVNet consists of a dual-stream radar backbone and a Radar Cross-Section (RCS) aware BEV encoder.
In the dual-stream radar backbone, a point-based encoder and a transformer-based encoder are proposed to extract radar features, with an injection and extraction module to facilitate communication between the two encoders.
The RCS-aware BEV encoder takes RCS as the object size prior to scattering the point feature in BEV.
Besides, we present the Cross-Attention Multi-layer Fusion module to automatically align the multi-modal BEV feature from radar and camera with the deformable attention mechanism, and then fuse the feature with channel and spatial fusion layers.
Experimental results show that \ours~achieves new state-of-the-art radar-camera fusion results on nuScenes and view-of-delft (VoD) 3D object detection benchmarks. 
Furthermore, \ours~achieves better 3D detection results than all real-time camera-only and radar-camera 3D object detectors with a faster inference speed at 21$\sim$28 FPS. 
The source code will be released at \url{https://github.com/VDIGPKU/RCBEVDet}.

\end{abstract}
\begin{figure}[!t]
    \centering
    \includegraphics[width=0.9\linewidth]{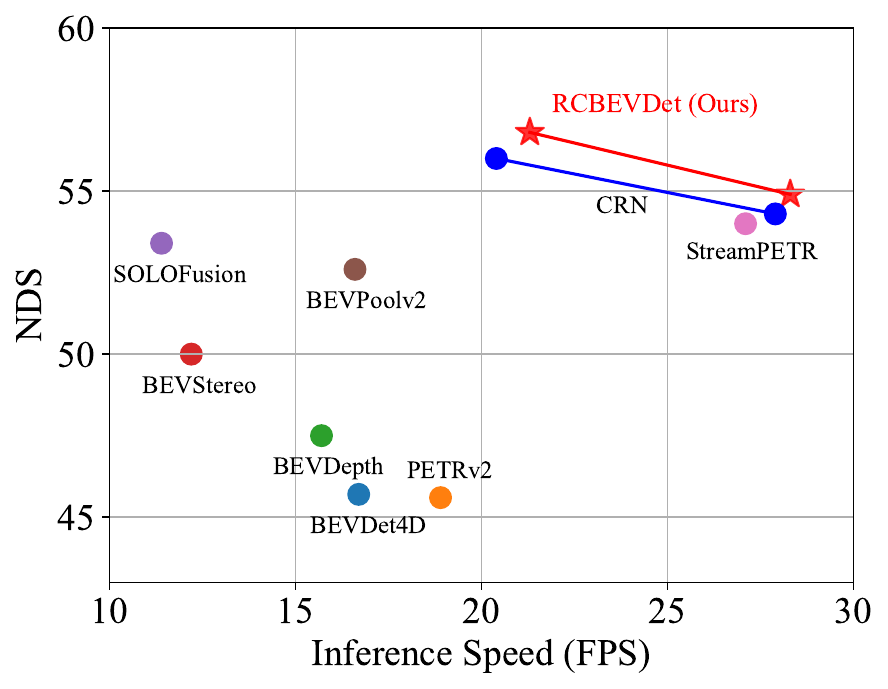}
    \vspace{-6pt}
    \caption{
    \textbf{Comparison of the proposed RCBEVDet and other real-time 3D object detectors.} Our RCBEVDet achieves state-of-the-art accuracy and accuracy-speed trade-offs. 
    All entries are evaluated on nuScenes \texttt{val} set, and the inference speed is benchmarked by a single RTX3090 GPU.
    }
    \vspace{-12pt}
    \label{fig:trade-off}
\end{figure}

\section{Introduction}
\label{sec:intro}
The rapid development of 3D object detection technology has greatly accelerated the progress of autonomous driving. 
Recently, researchers have focused on 3D object detection via multi-view cameras, which is economical and efficient.
Specifically, multi-view cameras can capture intricate details like object color and texture, and provide high-resolution semantic information for the 3D object detection task.
However, relying solely on a single camera sensor cannot achieve highly accurate and robust 3D object detection.
For instance, cameras cannot capture precise depth information~\cite{ma2021delving} and may fail in adverse weather or low-light conditions~\cite{bijelic2020seeing,DBLP:journals/corr/bevfusionmit}.

To overcome this issue, combining multi-view cameras with the economical millimeter-wave radar sensor to achieve more comprehensive and reliable multi-modal object detection is a feasible and effective solution.
Millimeter wave radar sensors excel in high-precision distance measurement and velocity estimation~\cite{charvat2014small}, and work reliably in various weather and lighting conditions~\cite{zhou2022towards,li2022exploiting}. 
Thus, millimeter wave radar sensors can provide complementary information to multi-view cameras.

Recent works~\cite{zheng2023rcfusion,xiong2023lxl} attempt to fuse radar and image at the feature level, and most of them follow the pipeline of BEVFusion~\cite{DBLP:journals/corr/bevfusion, DBLP:journals/corr/bevfusionmit} by projecting multi-view image features and radar features into bird's eye view (BEV).
Specifically, they mainly adopt simple element-wise concatenation, summation, or SE-module as the multi-modal feature fusion module.
However, these fusion methods cannot deal with the spatial misalignment between multi-view image and radar inputs.
Besides, current radar-camera fusion methods~\cite{kim2023craft, Kim_2023_ICCVCRN,nabati2021centerfusion} still adopt encoders (\textit{e.g.}, PointPillars) designed for the LiDAR sensor to process radar data.
Due to the natural differences between radar and LiDAR sensors, the LiDAR encoder adopted for radar data is sub-optimal.

To this end, we present a radar-camera 3D object detector dubbed \ours, which contains two key designs, \textit{i.e.}, RadarBEVNet for efficient radar feature extraction, and Cross-Attention Multi-layer Fusion Module (CAMF) for robust radar-camera feature fusion.
Specifically, RadarBEVNet consists of two components, \textit{i.e.}, the dual-stream radar backbone and the RCS-aware BEV encoder.
The first component, \textit{i.e.}, the dual-stream radar encoder, combines point-based and transformer-based encoders with an injection and extraction module.
To be more specific, the point-based encoder utilizes MLP to process each radar point individually, while the transformer-based encoder updates radar point features by interacting with other radar points.
Besides, the injection and extraction module is proposed as a connection for the interaction between features from two encoders.
In the second component, \textit{i.e.}, the RCS-aware BEV encoder, we consider RCS as a priori of object size to scatter point features into BEV space. 
As for the CAMF module, since radar points often suffer from azimuth errors, we first employ a multi-modal cross-attention mechanism to dynamically align the BEV feature maps from radar and cameras.
After the feature alignment, the channel and spatial fusion layers are applied to fuse the multi-modal features adaptively.

The main contributions of this work are summarized as follows:
\begin{compactitem}
    \item 
    We present \ours, a radar-camera multi-modal 3D object detector for highly accurate, efficient, and robust 3D object detection.
    \item 
    We specially design an efficient radar feature extractor for RCBEVDet, \textit{i.e.}, RadarBEVNet, consisting of a dual-stream radar backbone to extract radar features with two representations and an RCS-aware BEV encoder to scatter the radar feature into BEV according to radar-specific RCS character.
    
    \item
    We introduce the Cross-Attention Multi-layer Fusion module with the deformable cross-attention mechanism for robust radar-camera feature alignment and fusion.

    \item 
    RCBEVDet achieves new state-of-the-art radar-camera multi-modal 3D object detection results on nuScenes and VoD. 
    Besides, \ours~significantly improves the performance of camera-based 3D object detection methods and obtains optimal trade-off between accuracy and speed among all real-time camera-only and radar-camera 3D object detectors, as shown in \figref{fig:trade-off}.
    Furthermore, \ours~shows good robustness capability against sensor failure cases.

\end{compactitem}

\section{Related Work}
\label{sec:related-work} 

\subsection{Camera-based 3D Object Detection}

Detecting objects in 3D space from camera images is challenging due to the lack of sufficient 3D information compared to LiDAR and radar systems.
In recent years, researchers have made significant efforts to address this challenge~\cite{Graph-DETR3D,huang2021bevdet,li2022bevformer,shi2020points,weng2019monocular,chen2020monopair}.
These efforts include inferring depth from images~\cite{xu2021monocular}, utilizing geometric constraints and shape priors~\cite{lu2021geometry}, design specific loss functions~\cite{simonelli2020towards,miao2021pvgnet}, and exploring joint 3D detection and reconstruction optimization~\cite{liu2021voxel}.
More recently, the emergence of multi-view camera datasets~\cite{nuscenes,sun2020scalabilityWaymo} has led to the development of multi-view 3D object detection methods~\cite{philion2020liftLSS,huang2021bevdet,li2022bevformer,wang2021fcos3d,liu2022petr}, which can be briefly divided into two categories, \textit{i.e.}, geometry-based methods and transformer-based methods.

\begin{figure*}[!ht]
    \centering
    \includegraphics[width=0.95\linewidth]{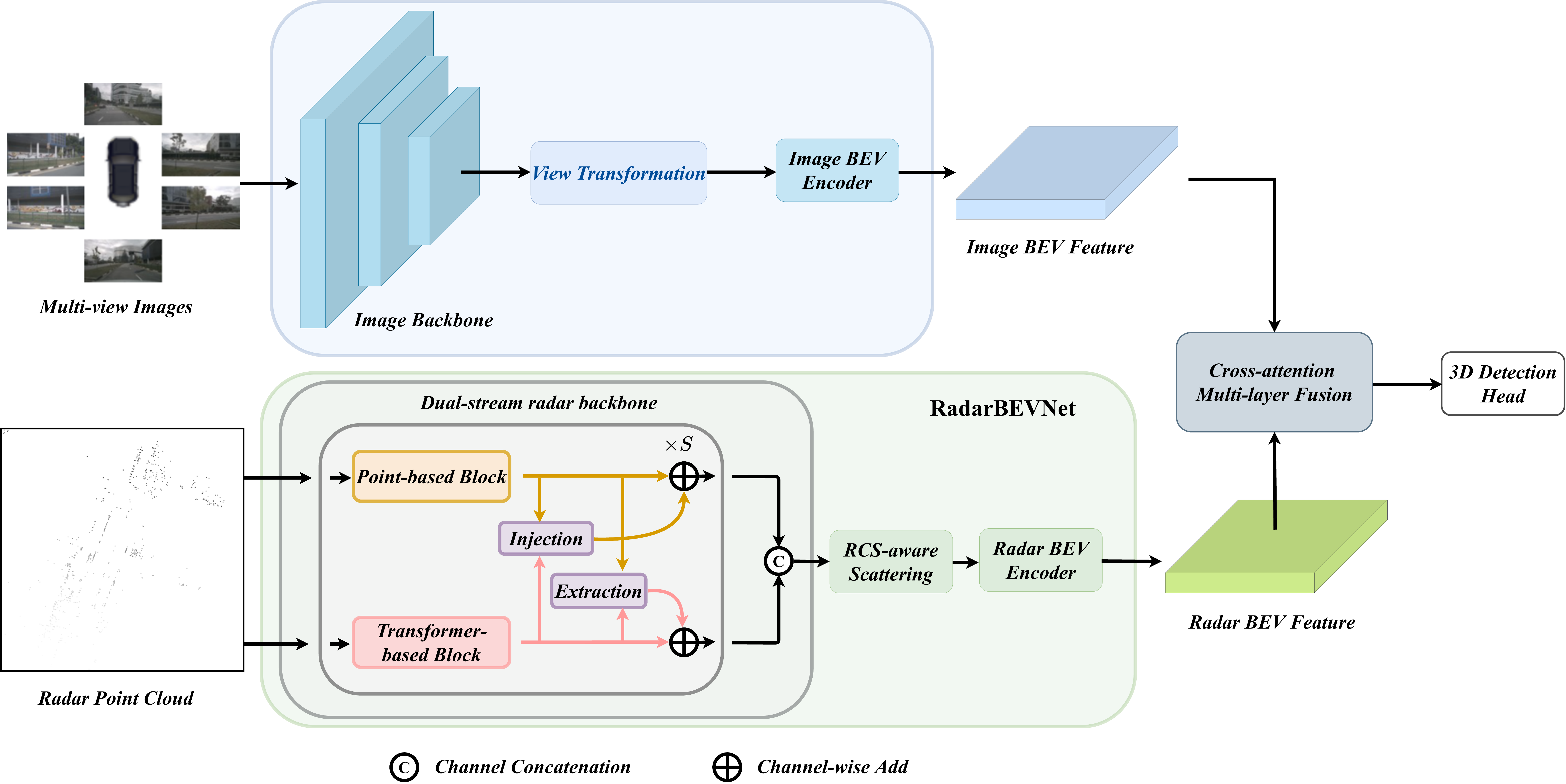}
    \caption{\textbf{Overall pipeline of \ours.}
    Firstly, multi-view images are encoded and transformed into the bird’s eye view to obtain the image BEV feature.
    Concurrently, radar point clouds are sent to the proposed RadarBEVNet to extract the radar BEV feature.
     Afterward, BEV features from radar and cameras are aligned dynamically and aggregated with the cross-attention multi-layer fusion (CAMF).
    The fused semantically rich multi-modal BEV feature is employed for the 3D object detection task.
    }
    \label{fig:pipeline}
    \vspace{-8pt}
\end{figure*}

Geometry-based methods mainly adopt Lift-Splat-Shoot (LSS)~\cite{philion2020liftLSS} to transform features from multi-view images into voxel or BEV features.
LSS~\cite{philion2020liftLSS} employs a depth estimation network to explicitly estimate the depth distribution and a context vector for each image. The outer product of the depth and context vector determines the feature at each point in 3D space along the perspective ray.
BEVDet~\cite{huang2021bevdet} builds upon the viewpoint transformation in LSS and detects 3D objects in BEV features.
BEVDepth~\cite{li2023bevdepth} introduces explicit depth supervision to optimize depth estimation. 
Based on BEVDet, BEVDet4D~\cite{huang2022bevdet4d} performs spatial alignment for BEV features from historical image frames and significantly improves the performance of velocity prediction.

As for Transformer-based methods, they mainly map the perspective view to a voxel or bird’s eye view by constructing queries and searching the corresponding image features through attention mechanisms.
BEVformer~\cite{li2022bevformer} introduces multi-scale deformable attention to map multi-view image features to BEV features.
StreamPETR~\cite{wang2023exploringStreamPETR} proposes to utilize sparse object queries as intermediate representations to capture temporal information.
SparseBEV~\cite{Liu_2023_ICCVSparseBEV} incorporates a scale-adaptive self-attention module and an adaptive spatio-temporal sampling module to perceive the BEV and temporal information dynamically.

Our proposed \ours~demonstrates that introducing additional complementary radar modality can significantly improve the 3D object detection performance while maintaining the real-time inference speed.

\subsection{Radar-camera 3D Object Detection}

Millimeter-wave radar is a popular sensor in autonomous vehicles for 3D object detection due to its cheapness, long-range perception, and Doppler velocity measurements, which are unaffected by adverse weather conditions. 
However, the sparsity and lack of semantic information of the millimeter-wave radar data make radar-only 3D object detection challenging. 
As a result, millimeter-wave radar is often adopted as the auxiliary modality for multi-modal 3D object detection.
Recently, combining multi-view cameras with millimeter-wave radar sensors for 3D object detection has attracted much attention, since millimeter-wave radar and multi-view images can provide complementary information for each other.

RadarNet~\cite{yang2020radarnet} proposes a multi-level fusion approach to improve the accuracy of distant objects and reduce velocity errors.
CenterFusion~\cite{nabati2021centerfusion} utilizes a keypoint detection network to generate initial 3D detection results from images and proposes a pillar-based radar association module to associate radar features with corresponding detection results for secondary refinement.
CRAFT~\cite{kim2023craft} introduces a proposal-level fusion framework that employs a Soft-Polar-Association and Spatio-Contextual Fusion Transformer to exchange information between the camera and millimeter-wave radar efficiently.
RADIANT~\cite{long2023radiant} designs a network for estimating the positional offset between radar echoes and object centers and leverages radar depth information to enhance camera features.
CRN~\cite{Kim_2023_ICCVCRN} leverages the depth information from radar to generate radar occupancy augmented images for multi-view transformation. It incorporates a cross-attention mechanism in radar-camera fusion to address spatial misalignment and information disparity between radar and camera sensors.
RCFusion~\cite{zheng2023rcfusion} designs radar PillarNet to generate radar pseudo-images and use a weighted fusion module to fuse radar and camera BEV features.

By contrast, our RCBEVDet proposes a well-designed RadarBEVNet for efficient radar BEV feature extraction and introduces a Cross-Attention Multi-layer fusion module for robust multi-modal feature alignment and fusion.

\section{Method}
\label{sec:method}
The overall pipeline of \ours~is shown in \figref{fig:pipeline}. 
Multi-view images are sent to an image encoder to extract features. Then, a view-transformation module is applied to transform the multi-view image feature into the image BEV feature.
Concurrently, aligned radar point clouds are encoded to the radar BEV feature by the proposed RadarBEVNet.
Afterward, the image and radar BEV features are fused by Cross-attention
Multi-layer Fusion module.
Finally, the fused multi-modal BEV feature is used for the 3D object detection task.

\subsection{RadarBEVNet}
\label{Sec:dual-stream encoder}
Previous radar-camera fusion methods mainly adopt the radar encoder designed for LiDAR point clouds, like PointPillars~\cite{lang2019pointpillars}.
Instead, we propose RadarBEVNet, especially for the efficient radar BEV feature extraction.

\begin{figure}
    \centering
    \begin{subfigure}{0.46\linewidth}
        \includegraphics[width=0.98\linewidth]{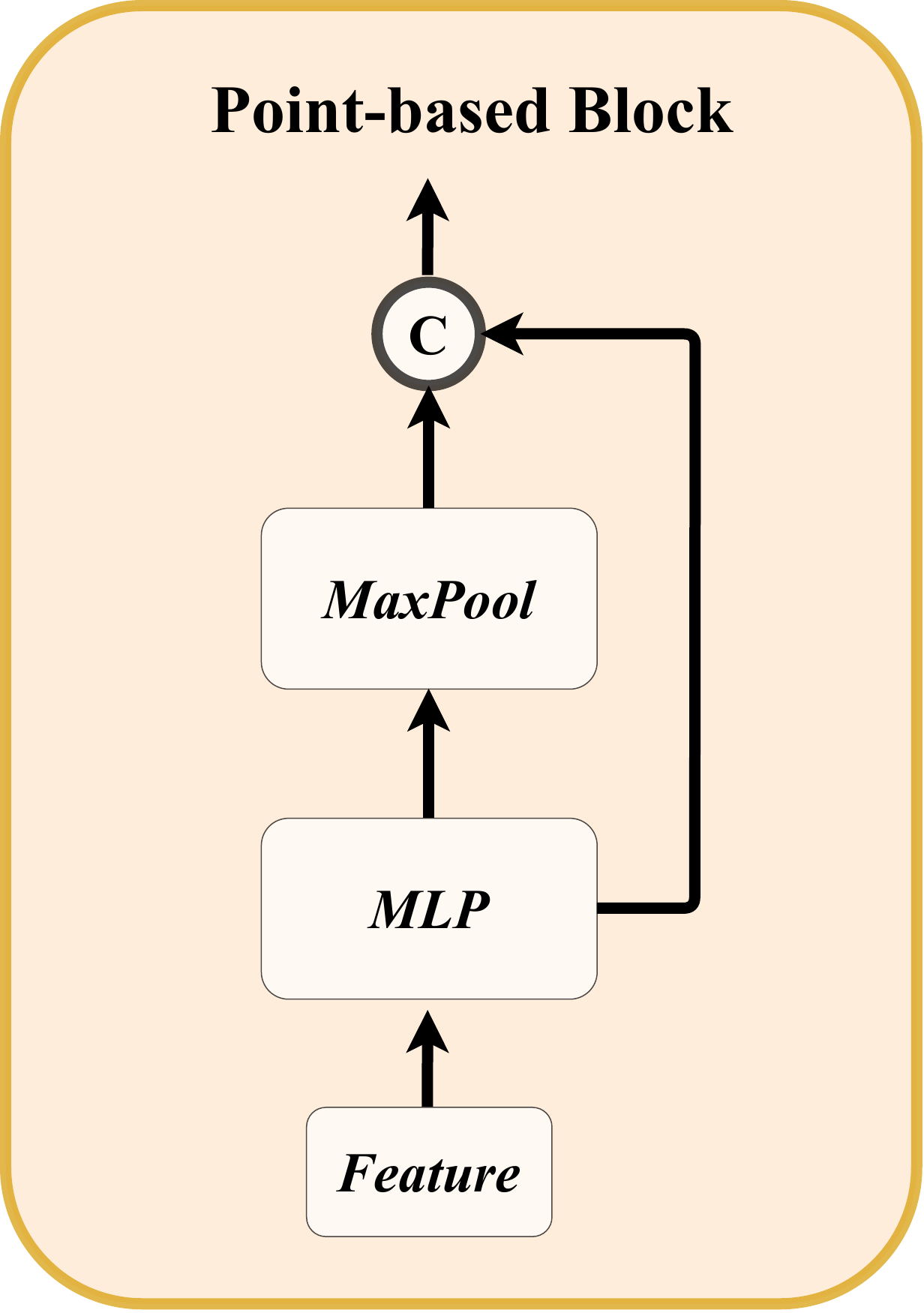}
        \caption{Point-based Block}
        \label{fig:point block}
    \end{subfigure}
    \quad
    \begin{subfigure}{0.46\linewidth}
        \includegraphics[width=0.98\linewidth]{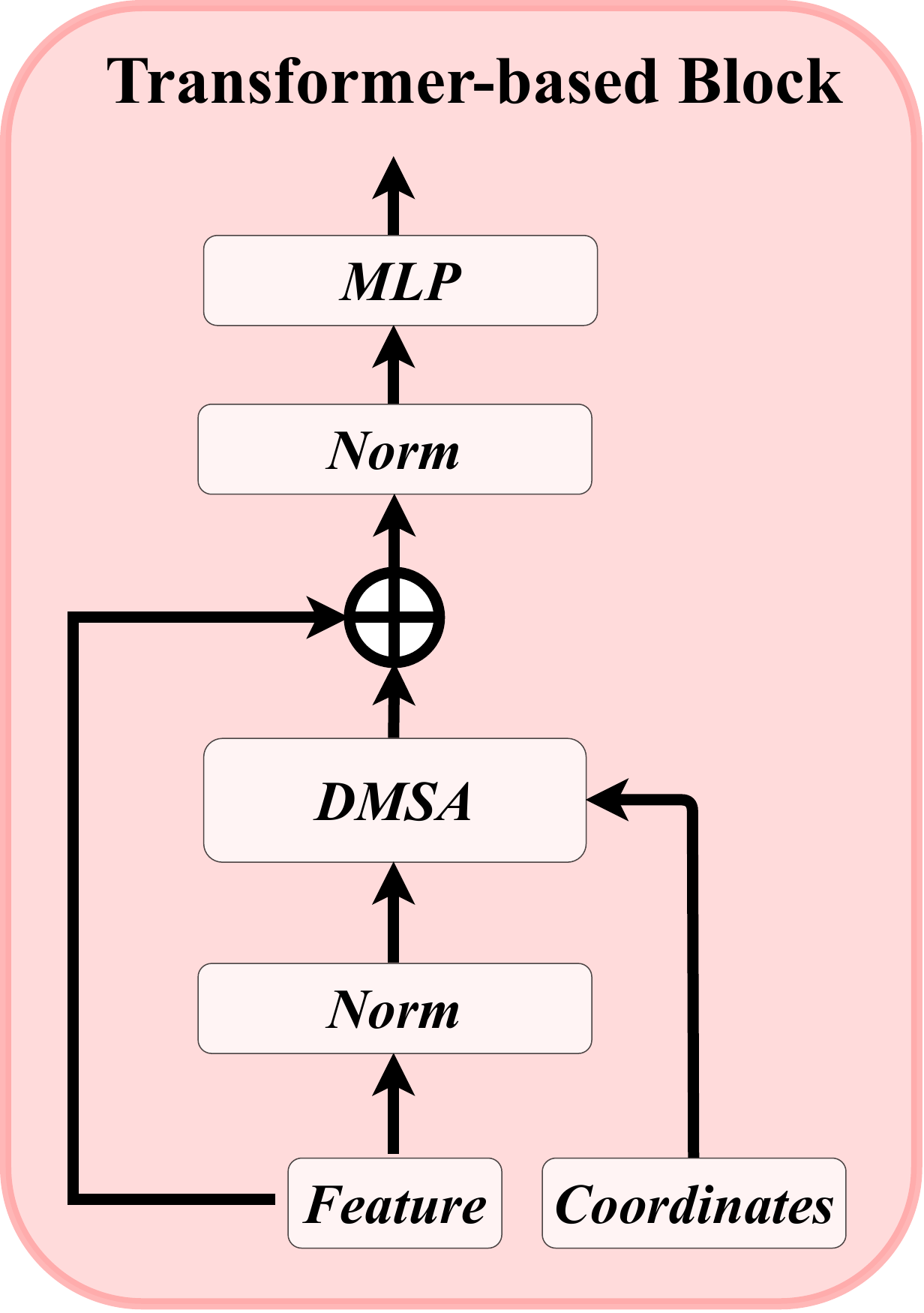}
        \caption{Transformer-based Block}
        \label{fig:transformer block}
    \end{subfigure}
    \caption{\textbf{Architecture of the dual-stream radar backbone.}}
    \label{fig:dual stream backbone}
    \vspace{-8pt}
\end{figure}

\noindent \textbf{Dual-stream radar backbone}.
The dual-stream radar backbone has two backbones, \textit{i.e.}, point-based backbone and transformer-based backbone.
The point-based backbone learns local radar features, while the transformer-based backbone captures global information.
Specifically, for the point-based backbone, we adopt a similar plain structure following PointNet~\cite{qi2017pointnet}.
As shown in \figref{fig:point block}, the point-based backbone has S blocks, and each block contains an MLP and a max pooling operation.
The input radar point feature is first sent to the MLP to increase its feature dimension.
Then, the global information is extracted by the max pooling operation over all radar points and concatenated to the high-dimension radar feature.
The whole process can be formulated as:
\vspace{-2pt}
\begin{equation}
    f = \text{Concat}[\text{MLP}(f),\text{MaxPool}(\text{MLP}(f))].
\end{equation}

\vspace{-3pt}
As for the transformer-based backbone, it comprises $S$ standard transformer blocks~\cite{vaswani2017attention, dosovitskiy2020vit} with an attention mechanism, a feed-forward network, and normalization layers, as shown in \figref{fig:transformer block}.
Due to the extensive range of autonomous driving scenarios, directly using standard self-attention can make it challenging to optimize the model.
To alleviate this issue, we propose a distance-modulated self-attention mechanism (DMSA) to make the model aggregate neighbor information at the early training iteration thus facilitating model convergence.
More specifically, given the coordinates of $N$ radar points, we first calculate the pair-distance $D\in \mathbb{R}^{N\times N}$ between all points. 
Then, we generate the Gaussian-like weight map $G$ according to the pair-distance $D$ as:
\vspace{-3pt}
\begin{equation}
    G_{i,j} = \exp(-D_{i,j}^2/\sigma^2),
    \vspace{-3pt}
\end{equation}
where $\sigma$ is a learnable parameter to control the bandwidth of the Gaussian-like distribution.
Essentially, the Gaussian-like weight map $G$ assigns high weight to spatial locations near the point and low weight to positions far from the point.
We modulate the attention mechanism with the generated weight $G$ as follows:
\vspace{-6pt}
\begin{equation}
\begin{split}
    \text{DMSA}(Q, K, V) = &\text{Softmax}(\frac{QK^{\top}}{\sqrt{d}}+\log G)V \\
    =&\text{Softmax}(\frac{QK^{\top}}{\sqrt{d}}-\frac{1}{\sigma^2}D^2)V.
\end{split}
\vspace{-3pt}
\end{equation}
To ensure DMSA can be degraded to vanilla self-attention, we replace $1/\sigma$ with a trainable parameter $\beta$ during the training.
When $\beta~\text{=}~0$, DMSA is degraded to the vanilla self-attention.
We also investigate the multi-head DMSA. Each head has unshared $\beta_i$ to control the receptive field of DMSA.
The multi-head DMSA can be formulated as $\text{MultiHeadDMSA}(Q, K, V)=\text{Concat}[head_1, head_2, ..., head_H ]$, where
\vspace{-2pt}
\begin{equation}
\begin{split}
    head_i &= DMSA(Q_i, K_i, V_i) \\
    &=\text{Softmax}(\frac{Q_iK_i^{\top}}{\sqrt{d_i}}-\beta_iD^2)V_i. \\
\end{split}
\end{equation}

\begin{figure}
    \centering
    \includegraphics[width=0.95\linewidth]{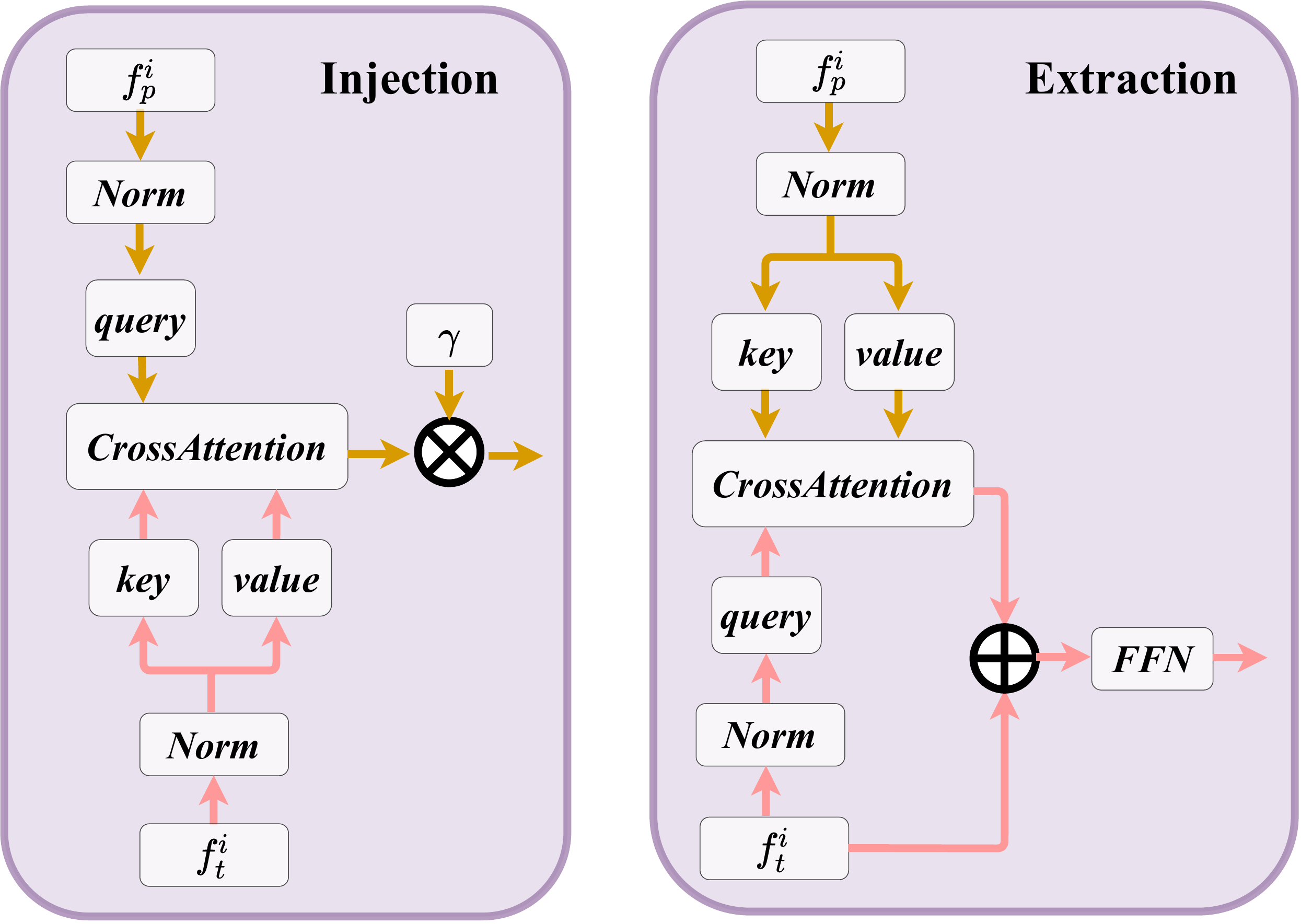}
    \caption{\textbf{Architecture of the Injection and Extraction module.} The left figure shows the details of the injection operation. The right figure displays the structure of the extraction operation.}
    \label{fig:injection_extraction}
    \vspace{-8pt}
\end{figure}

\vspace{-3pt}
To better interact radar features from two different backbones, we introduce the Injection and Extraction module based on cross-attention, as shown in \figref{fig:injection_extraction}.
The Injection and Extraction module is applied at each block of the two backbones.
Concretely, assuming the features from the $i~th$ block of the point-based and transformer-based backbone are $f_p^i$ and $f_t^i$, respectively.
In injection operation, we take $f_p^i$ as the query and $f_t^i$ as the key and value.
We use multi-head cross-attention to inject transformer feature $f_t^i$ into the point feature $f_p^i$, which can be formulated as:
\vspace{-2pt}
\begin{equation}
    f_p^i = f_p^i + \gamma\times \text{CrossAttention}(LN(f_p^i), LN(f_t^i)), \label{eq:injection}
    \vspace{-2pt}
\end{equation}
where \textit{LN} is the LayerNorm and $\gamma$ is a learnable scaling parameter.

Similarly, the extraction operation extract point feature $f_p^i$ with cross-attention for the transformer-based backbone.
The extraction operation is defined as:
\vspace{-2pt}
\begin{equation}
    f_t^i = \text{FFN}(f_t^i + \text{CrossAttention}(LN(f_t^i), LN(f_p^i))), \label{eq:extraction}\vspace{-2pt}
\end{equation}
where FFN is the FeedForward Network.
The updated features $f_p^i$ and $f_t^i$ are sent to the next block of their corresponding backbone.

\begin{figure}
    \centering
    \includegraphics[width=0.95\linewidth]{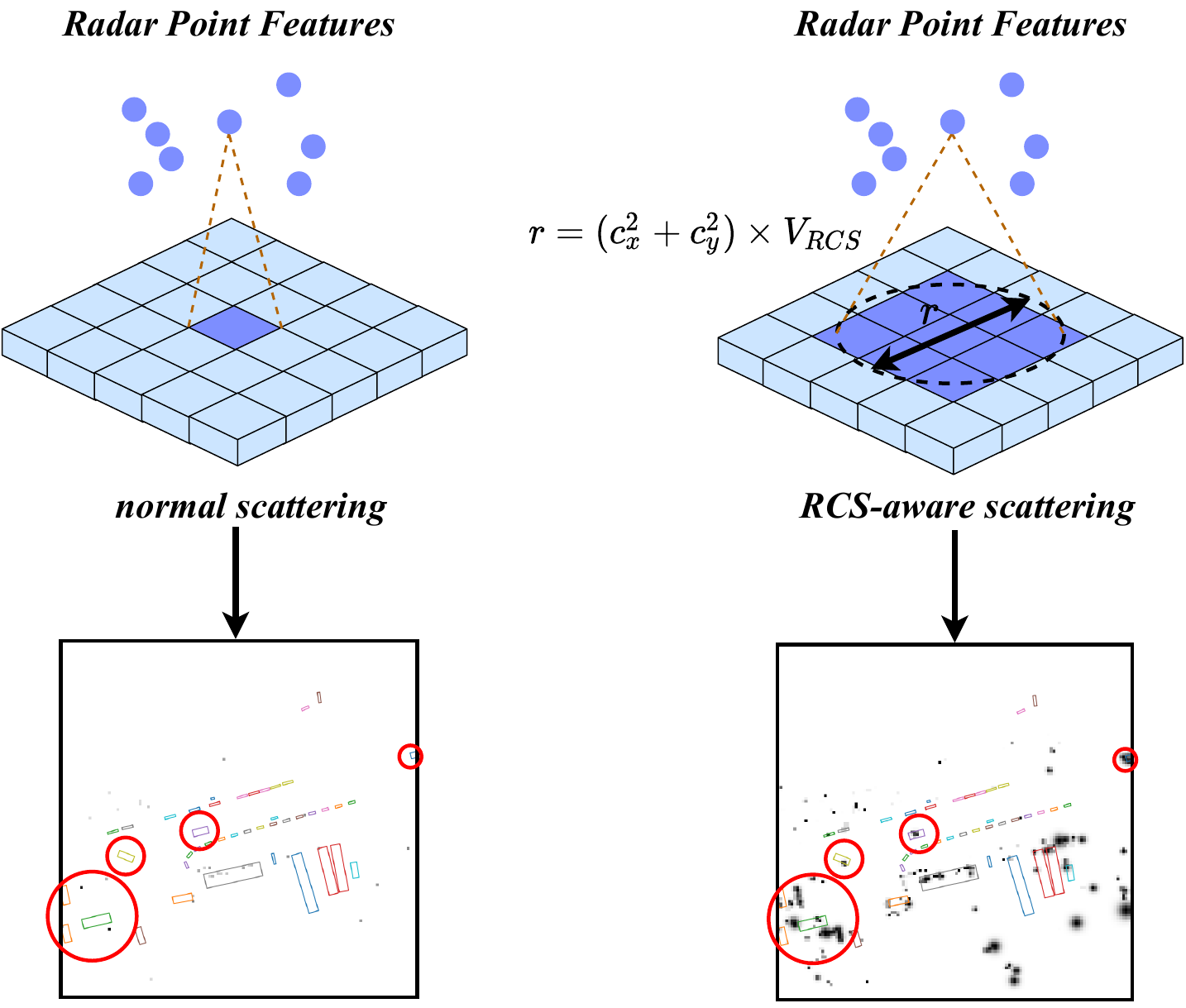}
    \caption{\textbf{Illustration of RCS-aware scattering.} RCS-aware scattering uses RCS as the object size prior to scatter the feature of one radar point to many BEV pixels. 
    }
    \label{fig:rcs}
    \vspace{-8pt}
\end{figure}

\vspace{3pt}
\noindent \textbf{RCS-aware BEV encoder}. 
Current radar BEV encoders generally scatter point features into voxel space according to the 3D coordinates of points and compress the z-axis to produce the BEV feature.
However, the produced BEV feature is sparse, \textit{i.e.}, the feature in most pixels is zero.
It is difficult for some pixels to aggregate features, which may hurt the detection performance.
One solution is to increase the number of BEV encoder layers.
This usually results in the features of small objects being smoothed out by background features.
To alleviate this problem, we propose an RCS-aware BEV encoder.
Radar cross-section (RCS) measures how an object is detectable by radar. 
In general, a larger object would produce stronger radar wave reflections, resulting in a larger RCS measurement.
Thus, RCS can provide a rough measurement of the object's size.
The key design of the RCS-aware BEV encoder is RCS-aware scattering operation, which utilizes RCS as the object size prior to scatter the feature of one radar point to many pixels instead of one pixel in BEV space, as shown in~\figref{fig:rcs}.

In particular, without loss of generality, given a specific radar point and its RCS value $v_{RCS}$, 3D coordinate $c=(c_x,c_y)$, BEV pixel coordinate $p=(p_x,p_y)$, and feature $f$, we scatter $f$ to pixel $p$ and nearby pixels, whose pixel distance with $p$ is smaller than $(c_x^2+c_y^2)\times v_{RCS}$.
If a pixel in the BEV feature is scattered with more than one radar feature, we perform summation pooling to aggregate these features.
In this way, we obtain the radar BEV feature $f_{RCS}$.
Besides, we introduce a Gaussian-like BEV weight map for each point according to the RCS value as:
\vspace{-2pt}
\begin{equation}
    G_{x,y}=\exp(-\frac{(c_x-x)^2+(c_y-y)^2}{\frac{1}{3}(c_x^2+c_y^2)\times v_{RCS}}),\vspace{-2pt}
\end{equation}
where $x,y$ are the pixel coordinates.
The final Gaussian-like BEV weight map $G_{RCS}$ is obtained by maximization over all Gaussian-like BEV weight maps.
We concatenate $f_{RCS}$ with $G_{RCS}$ and send them to an MLP to get the final RCS-aware BEV feature as:
\vspace{-2pt}
\begin{equation}
    f_{RCS}'=\text{MLP}(\text{Concat}(f_{RCS},G_{RCS})).\vspace{-2pt}
\end{equation}
After that, $f_{RCS}'$ is concatenated with the original BEV feature and sent to the BEV encoder, \textit{e.g.,} SECOND~\cite{yan2018second}.

\label{Sec: BEV cross attention}
\begin{figure}
    \centering
    \includegraphics[width=0.95\linewidth]{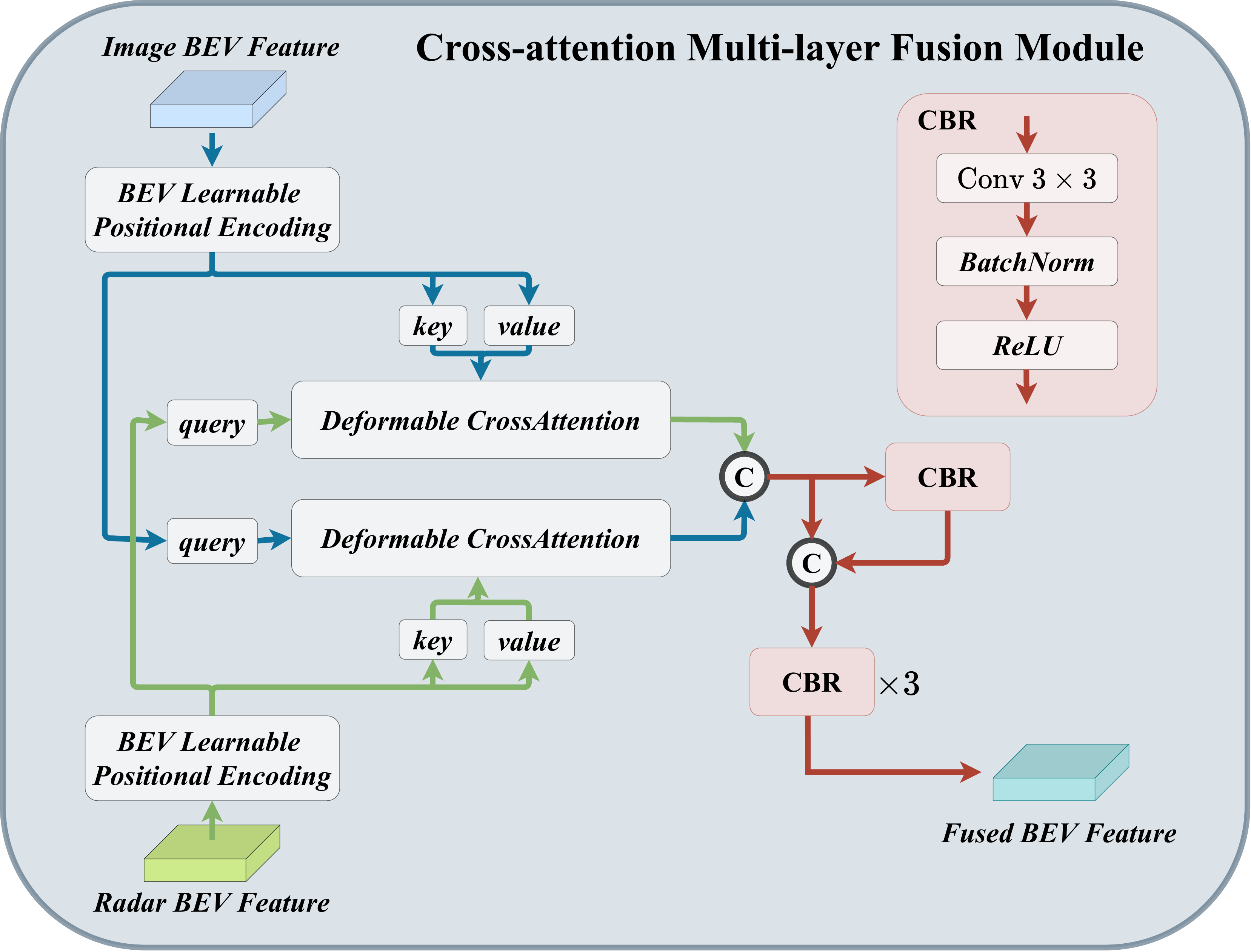}
    \caption{\textbf{Cross-attention Multi-layer Fusion module.} 
    The BEV features from radar and cameras are dynamically aligned with the deformable cross-attention.
    Then, the multi-modal BEV features are aggregated with a channel and spatial fusion module, which consists of several CBR blocks.
    }
    \label{fig:CAMF}
    \vspace{-5pt}
\end{figure}

\subsection{Cross-Attention Multi-layer Fusion Module}
\vspace{-2pt}

\setlength{\tabcolsep}{0.65em}
\begin{table*}[!t]
\caption{
    \textbf{Comparison of 3D object detection results on nuScenes \texttt{val} set}. 
    `C' and `R' represent camera and radar, respectively.
}
\vspace{-10pt}
\begin{center}
\label{table:3D det val set}
\resizebox{0.98\textwidth}{!}{
\begin{tabular}{l|c|c|c||cc|ccccc|c}
    \hline
    Method & Input & Backbone & Image Size & NDS$\uparrow$ & mAP$\uparrow$ & mATE$\downarrow$ & mASE$\downarrow$ & mAOE$\downarrow$ & mAVE$\downarrow$ & mAAE$\downarrow$ & FPS$\uparrow$ \\
    \hline
    CenterFusion \cite{nabati2021centerfusion}
    & C+R & DLA34 & $448\times800$  & 45.3 & 33.2 & 0.649 & \textbf{0.263} & 0.535 & 0.540 & \textbf{0.142} &-\\    
    CRAFT \cite{kim2023craft}
    & C+R & DLA34 & $448\times800$  & 51.7 & 41.1 & 0.494 & 0.276 & 0.454 & 0.486 & 0.176 &4.1 \\
    \gr \ours~(Ours) & C+R & DLA34 &  $448\times800$ & \textbf{56.3} & \textbf{45.3} & \textbf{0.492} & 0.269 & \textbf{0.449} & \textbf{0.230} & 0.188 & \textbf{4.7}\\
    \hline
    RCBEV4d \cite{RCBEV}
    & C+R & Swin-T & $256\times704$ & 49.7 & 38.1 & 0.526 & 0.272 & 0.445 & 0.465 & 0.185 &- \\
    \gr \ours~(Ours) & C+R & Swin-T &  $256\times704$ & \textbf{56.2} & \textbf{49.6}  & \textbf{0.496} &\textbf{0.271} & \textbf{0.418} & \textbf{0.239} & \textbf{0.179} &18.2\\
    \hline
    CRN~\cite{Kim_2023_ICCVCRN} & C+R& R18 & $256\times704$ & 54.3& \textbf{44.8}& 0.518& \textbf{0.283}& 0.552& 0.279& 0.180& 27.9 \\
    \gr \ours~(Ours) & C+R & R18 &  $256\times704$ &  \textbf{54.8} & 42.9 & \textbf{0.502}& 0.291& \textbf{0.432}& \textbf{0.210}& \textbf{0.178}  &\textbf{28.3}\\
    \hline
    BEVDet \cite{huang2021bevdet}
    & C & R50 & $256\times704$& 39.2 & 31.2 & 0.691 & 0.272 & 0.523 & 0.909 & 0.247 &15.6\\
    BEVDepth \cite{li2023bevdepth}
    & C & R50 & $256\times704$ & 47.5 & 35.1 & 0.639 & \textbf{0.267} & 0.479 & 0.428 & 0.198  &11.6\\
    SOLOFusion \cite{DBLP:conf/iclr/solofusion}
    & C & R50 & $256\times704$ & 53.4 & 42.7 & 0.567 & 0.274 & {0.411} & {0.252} & \textbf{0.188} &11.4\\
    StreamPETR~\cite{wang2023exploringStreamPETR} & C & R50 & $256\times704$ & 54.0 & 43.2 & 0.581 & 0.272 & 0.413 & 0.295 & 0.195 &\textbf{27.1}\\
    
    {CRN}~\cite{Kim_2023_ICCVCRN} 
    & C+R & R50 & $256\times704$&  {56.0} & \textbf{49.0} & {0.487} & 0.277 & 0.542 & 0.344 & 0.197 &20.4\\
    \gr \ours~(Ours) & C+R & R50 &  $256\times704$ & \textbf{56.8} & 45.3 & \textbf{0.486}  & 0.285 & \textbf{0.404} & \textbf{0.220} & 0.192 &21.3\\
    \hline
\end{tabular}}
\end{center}
\vspace{-11pt}
\end{table*}

\noindent\textbf{Multi-modal Feature Alignment with Cross-Attention}.
Radar point clouds often suffer from azimuth errors.
Thus, radar sensors may obtain radar points out of the boundaries of objects.
As a result, radar features generated by RadarBEVNet can be assigned to adjacent BEV grids, causing the misalignment of BEV features from cameras and radar.
To alleviate this issue, we use the cross-attention mechanism to align the multi-modal features dynamically.
Since unaligned radar points deviate from their true position by a small distance, we propose to employ the deformable cross-attention to capture the deviation.
Besides, the deformable cross-attention can reduce the computational complexity of the vanilla cross-attention from $O(H^{2}W^{2}C)$ to $O(HWC^2K)$, where $H$ and $W$ represent the height and width of the BEV feature, $C$ denotes the BEV feature channels, and $K$ is the number of the reference points in deformable cross-attention.

Specifically, as shown in \figref{fig:CAMF}, given camera and radar BEV features denoted by $ F_{c} \in \mathbb{R}^{C_{c} \times H \times W}$, $F_{r} \in \mathbb{R}^{C_{r} \times H \times W}$, respectively, we first add learnable position embeddings to $F_{c}$ and $F_{r}$.
Then, $F_{r}$ is transformed to queries $z_{q_r}$ and reference points $p_{q_r}$, and $F_{c}$ is viewed as keys and values.
The multi-head deformable cross-attention~\cite{zhu2020deformable} is calculated by:
\vspace{-3pt}
\begin{equation}
    \begin{aligned}
    &\operatorname{DeformAttn}\left(z_{q_r}, p_{q_r}, F_{c}\right)= \\
    &\sum_{m=1}^{M} \mathrm{~W}_{m}\left[\sum_{k=1}^{K} A_{m q k} \cdot \mathrm{W}_{m}^{\prime} F_c\left(p_{q_r}+\Delta p_{m q k}\right)\right],\vspace{-3pt}
\end{aligned}
\end{equation}
where $m$ indexes the attention head, $k$ indexes the sampled keys, and $K$ is the total sampled key number. $\Delta p_{m q k}$ denotes the sampling offset. $A_{mqk}$ represents attention weight calculated by $z_{q_r}$ and $F_c$. 
We exchange $F_{r}$ and $F_{c}$ and conduct another deformable cross-attention to update $F_{r}$.

Finally, the Cross-Attention module in CAMF can be formulated as:
\begin{equation}
\vspace{-3pt}
\left\{\begin{aligned}
F_{c} & \leftarrow \operatorname{DeformAttn}\left(z_{q_{r}}, p_{q_{r}}, F_{c}\right), \\
F_{r} & \leftarrow \operatorname{DeformAttn}\left(z_{q_{c}}, p_{q_{c}}, F_{r}\right). \\
\end{aligned}\right.
\vspace{-3pt}
\end{equation}


\label{Sec: BEV fusion}
\vspace{3pt}
\noindent\textbf{Channel and Spatial Fusion}.
After aligning the BEV feature from cameras and radar by cross-attention, we propose channel and spatial fusion layers to aggregate multi-modal BEV features, as illustrated in ~\figref{fig:CAMF}.
Specifically, we first concatenate two BEV features as ${F}_{multi}=\left[{F}_{{c}}, {F}_{{r}}\right]$.
Then, ${F}_{multi}$ is sent to a CBR block with a residual connection to obtain the fused feature. 
The CBR block is successively composed of a Conv $3\times3$, a Batch Normalization, and a ReLU activate function.
After that, three CBR blocks are applied to further fuse the multi-modal features.

\renewcommand{\thefootnote}{1}
\setlength{\tabcolsep}{0.65em}
\begin{table*}[!t]
\begin{center}
\caption{
    \textbf{Comparison of 3D object detection results on nuScenes \texttt{test} set}.
    `C' and `R' represent camera and radar, respectively.
    $^{*}$ indicates the camera model in RCBEVDet, which is implemented based on the released official code.
}
\vspace{-5pt}
\label{table:3D det test set}
\resizebox{0.94\textwidth}{!}{
\begin{tabular}{l|c|c||cc|ccccc}
    \hline
    Method & Input  & Backbone & NDS$\uparrow$ & mAP$\uparrow$ & mATE$\downarrow$ & mASE$\downarrow$ & mAOE$\downarrow$ & mAVE$\downarrow$ & mAAE$\downarrow$\\
    \hline
    \small{KPConvPillars} \cite{KPConvPillars}   & R & Pillars    & 13.9 &  4.9 & 0.823 & 0.428 &0.607&2.081&1.000 \\
    \hline
    CenterFusion \cite{nabati2021centerfusion}  &C+R& DLA34    & 44.9 & 32.6 & 0.631 & 0.261 & 0.516 &0.614&0.115 \\
    
    RCBEV \cite{RCBEV}               &C+R& Swin-T   & 48.6 & 40.6 & 0.484 & 0.257 & 0.587& 0.702 &0.140 \\
    
    
    MVFusion \cite{DBLP:conf/icra/mvfusion}              &C+R& V2-99    & 51.7 & 45.3 & 0.569 &  0.246 & 0.379& 0.781& 0.128 \\
    
    CRAFT \cite{kim2023craft}                   &C+R& DLA34    & 52.3 & 41.1 & 0.467 & 0.268 & 0.456 &0.519&0.114 \\
    
    BEVFormer \cite{li2022bevformer}            & C & V2-99    & 56.9 & 48.1 & 0.582 & 0.256 &0.375& 0.378& 0.126 \\
    PETRv2~\cite{liu2023petrv2} & C & V2-99 & 58.2 & 49.0 & 0.561& 0.243& 0.361& 0.343& 0.120 \\
    BEVDepth$^{*}$ \cite{li2023bevdepth}              & C & V2-99 & 60.5 & 51.5 & 0.446&  0.242& 0.377& 0.324& 0.135 \\
    BEVDepth \cite{li2023bevdepth}              & C & \small{ConvNeXt-B} & 60.9 & 52.0 & 0.445&  0.243& 0.352& 0.347& 0.127 \\
    
    BEVStereo \cite{li2019stereo}            & C & V2-99    & 61.0 & 52.5 & 0.431 & 0.246 &0.358& 0.357& 0.138 \\
    
    SOLOFusion \cite{DBLP:conf/iclr/solofusion}              & C & \small{ConvNeXt-B} & 61.9 & 54.0 & 0.453 & 0.257& 0.376 &0.276& 0.148 \\
    
    {CRN}~\cite{Kim_2023_ICCVCRN}                            &C+R& \small{ConvNeXt-B} & {62.4} & \textbf{57.5} & 0.416 & 0.264 & 0.456 & 0.365 & 0.130 \\
    SparseBEV~\cite{Liu_2023_ICCVSparseBEV} & C & V2-99& 63.6 &55.6 &0.485& 0.244& \textbf{0.332}& 0.246& 0.117\\
    {StreamPETR}~\cite{wang2023exploringStreamPETR}                            &C& {V2-99} &63.6 & 55.0  &0.493& 0.241 &{0.343}& \textbf{0.243}& 0.123 \\
    \gr \ours~(Ours)                          &C+R& {V2-99} & \textbf{63.9} & 55.0 & \textbf{0.390} &\textbf{0.234} & 0.362 & 0.259 & \textbf{0.113} \\
    \hline
\end{tabular}}
\end{center}
\vspace{-5pt}
\end{table*}

\section{Experiments}
\subsection{Implementation Details.}
We adopt BEVDepth~\cite{li2023bevdepth} with several modifications as the camera stream in \ours. 
We accumulate the intermediate BEV feature of the previous frames and concatenate them with the BEV feature of the current frame.
We follow BEVDet4D~\cite{huang2022bevdet4d} to add an extra BEV encoder consisting of two residual blocks to adjust the BEV features from multi-frames before the temporal fusion.
We use BEVPoolv2~\cite{huang2022bevpoolv2} to accelerate the view transformation process.
For radar, we accumulate multi-sweep radar points and use RCS and Doppler speed as features in the same manner as GRIFNet~\cite{kim2020grif} and CRN~\cite{Kim_2023_ICCVCRN}. 
The number of stage $S$ in the dual-stream radar backbone is set to three.

Our models are trained in a two-stage manner.
In the first stage, we train the camera stream following the standard implementation~\cite{li2023bevdepth}.
In the second stage, we train the radar-camera fusion model.
The weights of the camera stream are inherited from the first stage, and the parameters of the camera stream are frozen in the second stage.
All models are trained for 12 epochs with AdamW~\cite{kingma2015adam} optimizer.
We apply both image and radar data augmentations to prevent overfitting.
We adopt CBGS~\cite{zhu2019cbgs} for the class-balanced sampling.
Inference time is measured on an RTX3090 GPU with a single batch and FP16 precision following CRN~\cite{Kim_2023_ICCVCRN}.

\renewcommand{\thefootnote}{1}
\setlength{\tabcolsep}{0.65em}
\begin{table*}[!t]
\begin{center}
\caption{
    \textbf{Comparison of 3D object detection results on VoD validation set}.
    The region of interest is the driving corridor located close to the ego-vehicle.
    The IoU thresholds for mAP~\cite{palffy2022multiVOD} are set to 0.5 for cars, 0.25 for pedestrians, and 0.25 for cyclists.
}
\vspace{-5pt}
\label{table:vod}
\resizebox{0.88\textwidth}{!}{
\begin{tabular}{l|c||ccc|c|ccc|c}
    \hline
    \multirow{2}{*}[-0.5ex]{Method} & \multirow{2}{*}[-0.5ex]{Input}  & \multicolumn{4}{c|}{AP in the Entire Annotated Area (\%)} & \multicolumn{4}{c}{AP in the Region of Interest (\%)} \\
    \cline{3-10}
    & & Car & Pedestrian &Cyclist& mAP & Car &Pedestrian& Cyclist& mAP \\
    \hline
    PointPillar~\cite{lang2019pointpillars} & R &37.06& 35.04& 63.44& 45.18& 70.15& 47.22& 85.07& 67.48 \\
    RadarPillarNet~\cite{zheng2023rcfusion} & R& 39.30& 35.10& 63.63& 46.01 &71.65& 42.80& 83.14& 65.86 \\
    RCFusion~\cite{zheng2023rcfusion} & C+R &41.70& 38.95& 68.31& 49.65& 71.87& 47.50& 88.33& 69.23 \\
    \gr \ours~(Ours) &  C+R & 40.63& 38.86 & 70.48 & \textbf{49.99} 
    & 72.48 & 49.89 & 87.01 & \textbf{69.80} \\ 
    \hline
\end{tabular}}
\end{center}
\vspace{-15pt}
\end{table*}

\subsection{Main Results}

\setlength{\tabcolsep}{0.6em}
\begin{table}[!t]
\caption{
\textbf{Ablation of the main components of \ours}. We successively add components to BEVDepth~\cite{li2023bevdepth} and compose \ours. Each component improves the 3D detection performance consistently.
}
\vspace{-15pt}
\definecolor{red}{HTML}{39b54a}  
\definecolor{green}{HTML}{ea4335}  
\begin{center}
\label{table:ablation}
\resizebox{0.98\columnwidth}{!}{
\begin{tabular}{l|c||cc}
\hline
Model Configuration & Input & NDS & mAP\\
\hline
BEVDepth~\cite{li2023bevdepth} & C & 47.5 & 35.1 \\ 
+ Temporal  & C & 51.9 \textcolor{green}{$\uparrow$4.4} & 40.5 \textcolor{green}{$\uparrow$5.4} \\
\hline
+ PointPillar+BEVFusion  & C+R & 53.6 \textcolor{green}{$\uparrow$1.7} & 42.3 \textcolor{green}{$\uparrow$1.8} \\
+ RadarBEVNet  & C+R & 55.7 \textcolor{green}{$\uparrow$2.1} & 45.3 \textcolor{green}{$\uparrow$3.0} \\
+ CAMF  & C+R & 56.4 \textcolor{green}{$\uparrow$0.7} & \textbf{45.6} \textcolor{green}{$\uparrow$0.3} \\
+ Temporal Supervision  & C+R & \textbf{56.8} \textcolor{green}{$\uparrow$0.4} & 45.3 \textcolor{red}{$\downarrow$0.3} \\
\hline
\end{tabular}
}
\end{center}
\vspace{-21pt}
\end{table}

\noindent\textbf{NuScenes Results.}
We compare the proposed RCBEVDet with previous state-of-the-art 3D detection methods on the nuScenes \texttt{val} and \texttt{test} sets in Tables \ref{table:3D det val set} and \ref{table:3D det test set}, respectively.
As shown in \tabref{table:3D det val set}, \ours~shows competitive 3D object detection performance, especially on the overall metrics (NDS) and velocity error (mAVE).
Specifically, \ours~outperforms previous radar-camera fusion methods under various backbone settings with a faster inference speed.
Notably, \ours~with ResNet-50 reduces the velocity error (mAVE) by 14.7\% and 37.5\% compared to the previous best camera-only method (SOLOFusion) and radar-camera method (CRN), respectively.
Furthermore, \ours~surpasses all camera-based 3D detection methods, showing the effectiveness of using complementary radar information for better 3D detection.

On \texttt{test} sets, RCBEVDet improves the baseline (BEVDepth) results by 3.4 NDS and 3.5 mAP.
It is worth noting that we can further improve the performance of RCBEVDet by adopting a stronger detection baseline, such as StreamPETR~\cite{wang2023exploringStreamPETR}.
Besides, RCBEVDet with a smaller V2-99 backbone can surpass the radar-camera fusion method CRN with ConvNeXt-Base by 1.5 NDS.
%

\noindent\textbf{VoD Results.} 
To further demonstrate the effectiveness of RCBEVDet, we train RCBEVDet on the 4D millimeter-wave radar dataset, \textit{i.e.}, view-of-delft (VoD).
We report the results of \ours~on the VoD validation set in \tabref{table:vod}. 
%
In the entire area, RCBEVDet surpasses RCFusion by 0.34 mAP.
As for the region of interest, RCBEVDet also achieves state-of-the-art results with 69.80 mAP.
%


\subsection{Ablation Studies}
We conduct ablation studies on nuScenes \texttt{val} set to analyze the effectiveness of each setting of RCBEVDet. 
We adopt RCBEVDet with R50 backbone, $256\times704$ image size, and $ 128\times128$ BEV size as the baseline model.

\vspace{2pt}
\noindent\textbf{Main Components.}
RCBEVDet improves over the BEVDepth by combining it with several main components described in \secref{sec:method}. 
To evaluate the effectiveness of each component, we successively add components to
a baseline BEVDepth and compose \ours.
As shown in \tabref{table:ablation}, we observe that each component consistently improves performance.
For the camera-only model, adding temporal information by accumulating BEV features from multi-frames can bring 4.4 NDS and 5.4 mAP improvement.
Based on the camera-only model, we build a radar-camera 3D object detection baseline by adopting PointPillar as the radar backbone and BEVFusion~\cite{DBLP:journals/corr/bevfusion} as the fusion method, which uses a convolutional layer and SE~\cite{hu2018squeeze} to fuse multi-model BEV features.
The baseline radar-camera detector can achieve 53.6 NDS and 42.3 mAP.
By replacing PointPillar with the proposed RadarBEVNet, we obtain 2.1 NDS and 3.0 mAP improvement, showing the excellent radar feature representation ability of RadarBEVNet.
Moreover, the proposed CAMF can further improve the detection performance from 55.7 NDS to 56.4 NDS.
Additionally, following HoP~\cite{Zong_2023_hop}, we add extra losses to supervise the output of each frame, named Temporal Supervision. 
This can improve 0.4 NDS while decreasing 0.3 mAP.

Overall, the results demonstrate the effectiveness of each component proposed in \ours. 
Meanwhile, the results indicate that multi-module fusion is essential for the 3D object detection task.


\setlength{\tabcolsep}{0.6em}
\begin{table}[!t]
\caption{
\textbf{Ablation of RadarBEVNet}. 
}
\vspace{-14pt}
\definecolor{green}{HTML}{ea4335}  
\begin{center}
\label{table:RadarBEVNet}
\resizebox{0.98\columnwidth}{!}{
\begin{tabular}{l||cc}
\hline

Radar Backbone & NDS & mAP\\
\hline

Point Backbone & 54.3 & 42.6 \\ 
+ RCS-aware BEV encoder   & 55.7 \textcolor{green}{$\uparrow$1.4} & 44.5 \textcolor{green}{$\uparrow$1.9} \\
\hline
+ Transformer Backbone  & 55.8 \textcolor{green}{$\uparrow$0.1} & 44.8 \textcolor{green}{$\uparrow$0.3} \\
+ Injection and Extraction module  & \textbf{56.4} \textcolor{green}{$\uparrow$0.6} & \textbf{45.6} \textcolor{green}{$\uparrow$0.8} \\
\hline
\end{tabular}
}
\end{center}
\vspace{-10pt}
\end{table}

\setlength{\tabcolsep}{0.6em}
\begin{table}[!t]
\caption{
\textbf{Ablation of CAMF}. 
}
\vspace{-14pt}
\definecolor{green}{HTML}{ea4335}  
\begin{center}
\label{table:CAMF}
\resizebox{0.98\columnwidth}{!}{
\begin{tabular}{l||cc}
\hline

Fusion Method & NDS & mAP\\
\hline

BEVFusion~\cite{DBLP:journals/corr/bevfusionmit} & 55.7 & 45.3 \\ 
\hline
+ Deformable Cross-Attention  & 56.1 \textcolor{green}{$\uparrow$0.4} & 45.5 \textcolor{green}{$\uparrow$0.2} \\
+ Channel and Spatial Fusion  & \textbf{56.4} \textcolor{green}{$\uparrow$0.3} & \textbf{45.6} \textcolor{green}{$\uparrow$0.1} \\
\hline
\end{tabular}
}
\end{center}
\vspace{-18pt}
\end{table}

\vspace{2pt}
\noindent\textbf{RadarBEVNet.}
We conduct the ablation experiments for the design of RadarBEVNet, including the Dual-stream radar backbone and RCS-aware BEV encoder, as shown in \tabref{table:RadarBEVNet}.
The baseline model, which replaces RadarBEVNet with a single Point-based backbone, can achieve 54.3 NDS and 42.6 mAP.
By using the RCS-aware BEV encoder, we can improve the performance of 3D detection by 1.4 NDS and 1.9 mAP, showing the effectiveness of the proposed RCS-aware BEV feature reconstruction.
Besides, we observe that directly adding the Transformer-based Backbone brings marginal performance improvement.
The reason is that the Point-based backbone and the Transformer-based Backbone process radar points individually.
The different radar feature representations of two backbones cannot be merged well.
To overcome this issue, the Injection and Extraction module is introduced and brings performance gain with 0.6 NDS and 0.8 mAP.


\vspace{2pt}
\noindent\textbf{CAMF module.}
We conducted ablation experiments on the CAMF module, including the deformable cross-attention mechanism for multi-modal alignment and the channel and spatial fusion module, as shown in \tabref{table:CAMF}.
The baseline model with the fusion module in BEVfusion~\cite{DBLP:journals/corr/bevfusionmit} can achieve 55.7 NDS and 45.3 mAP.
Incorporating the Deformable Cross Attention for BEV features alignment yields 3D detection performance improvement from 55.7 NDS to 56.1 NDS and 45.3 mAP to 45.5 mAP. 
The results demonstrate the effectiveness of the cross-modal alignment capability of the cross-attention mechanism.
Besides, we observed that introducing the channel and spatial fusion module for BEV feature fusion resulted in a higher performance of 0.3 NDS and 0.1 mAP than one-layer fusion in BEVFusion~\cite{DBLP:journals/corr/bevfusionmit}.

\subsection{Analysis of Robustness}
\begin{table}[!t]
\caption{
\textbf{Analysis of Robustness}. We report \textit{Car} class mAP in the same manner as CRN~\cite{Kim_2023_ICCVCRN}.}
\vspace{-12pt}
\begin{center}
\label{table:robust}
\resizebox{0.94\columnwidth}{!}{
\begin{tabular}{l|c|c||cccc}
\hline
& \multirow{2}{*}{Input} & \multirow{2}{*}{Drop} & \multicolumn{4}{c}{\# of view drops} \\
&  &  & 0 & 1 & 3 & All \\
\hline
BEVDepth~\cite{li2023bevdepth}    &  C  & C & 49.4 & 41.1 & 24.2 & 0 \\
CenterPoint~\cite{yin2021centerpoint} &  R  & R & 30.6 & 25.3 & 14.9 & 0 \\
\hline
\multirow{2}{*}{BEVFusion~\cite{DBLP:journals/corr/bevfusionmit}} & \multirow{2}{*}{C+R} & C & \multirow{2}{*}{63.9} 
& 58.5 & 45.7 & {14.3} \\
& & R & & 59.9 & 50.9 & 34.4 \\
\hline

\multirow{2}{*}{CRN~\cite{Kim_2023_ICCVCRN} }      & \multirow{2}{*}{C+R} & C & \multirow{2}{*}{\shortstack{{68.8}}} 
& {62.4} & {48.9} & 12.8 \\

& & R & & {64.3} & {57.0} & {43.8} \\

\hline
\multirow{2}{*}{\ours~(Ours)}       & \multirow{2}{*}{C+R} & C & \multirow{2}{*}{\shortstack{\textbf{72.5}}} 
& \textbf{66.9} & \textbf{53.5} & \textbf{16.5} \\

& & R & & \textbf{71.6} & \textbf{66.1} & \textbf{62.1} \\

\hline
\end{tabular}}
\end{center}
\vspace{-18pt}
\end{table}

We randomly drop images or radar inputs to analyze the robustness in sensor failure cases.
We use the dropout training strategy as the data augmentation to train \ours~in this experiment following CRN~\cite{Kim_2023_ICCVCRN}.
The results with \textit{Car} class mAP are present in \tabref{table:robust}.
\ours~outperforms CRN and BEVFusion with higher \textit{Car} class mAP in every sensor failure case.
Notably, the performance of CRN~\cite{Kim_2023_ICCVCRN} decreases by 4.5, 11.8, and 25.0 mAP in three radar sensor failure cases, while \ours~only drops by 0.9, 6.4, and 10.4 mAP.
The results demonstrate that the proposed cross-attention module can produce a more robust BEV feature with the dynamic BEV feature alignment.

\section{Conclusion}
In this paper, we present \ours, a radar-camera multi-modal 3D object detector in the bird’s eye view.
We specially design RadarBEVNet for efficient radar BEV feature extraction with a dual-stream radar backbone and an RCS-aware BEV encoder.
Besides, we propose a CAMF module to dynamically align radar-camera BEV features and achieve robust 3D object detection.
Experimental results show that, with radar sensors, RCBEVDet obtains a significant performance gain over camera-based 3D object detection methods, while maintaining real-time inference speed.
Furthermore, RCBEVDet achieves new state-of-the-art radar-camera fusion results on nuScenes and VoD 3D object detection benchmarks. 


{
    \small
    \bibliographystyle{ieeenat_fullname}
    \bibliography{main}
}


\end{document}